\documentclass[letterpaper, 10 pt, conference]{ieeeconf}  

\IEEEoverridecommandlockouts                              
\usepackage[
backend=bibtex8, 
bibencoding=utf8,
style=ieee, 
sorting=none, 
natbib=true, 
doi=false, 
isbn=false, 
url=false, 
eprint=false, 
maxcitenames=1, 
mincitenames=1,
]{biblatex}

\renewbibmacro*{bbx:savehash}{}

\usepackage[draft,pdftex,colorlinks]{hyperref}

\usepackage[printonlyused]{acronym}

\usepackage{siunitx}
\usepackage[all]{nowidow}

\usepackage[dvipsnames]{xcolor}

\usepackage{lipsum}

\usepackage[pdftex]{graphicx}

\usepackage{epstopdf}

\usepackage{import}

\graphicspath{{./latexGoodPractices/}}

\usepackage{booktabs}

\usepackage{tabu}

\usepackage{amssymb,amsfonts,amsmath,amscd}

\usepackage{bm} 

\newcommand{\bbm}{\begin{bmatrix}}
\newcommand{\ebm}{\end{bmatrix}}

\usepackage{subcaption}
\usepackage{commath}

\newcommand{\floor}[1]{\lfloor #1 \rfloor}
\newcommand{\ie}{i.e., }
\newcommand{\eg}{e.g., }
\newcommand{\expp}[1]{\exp\left(#1\right)}
\newcommand{\pr}[1]{\mathbb{P}\left(#1\right)}
\newcommand{\area}{\mathcal{A}}

\title{\LARGE \bf
		Lambda-Field: A Continuous Counterpart of the Bayesian Occupancy Grid for Risk Assessment 
}

\author{Johann Laconte$^{1}$, Christophe Debain$^{3}$, Roland Chapuis$^{1}$, Fran\c cois Pomerleau$^{2}$, Romuald Aufr\`ere$^{1}$ 
\thanks{$^{1}$ Universit\'e Clermont Auvergne, CNRS, SIGMA Clermont, Institut Pascal, F-63000 CLERMONT-FERRAND, FRANCE; johann.laconte@uca.fr}%
\thanks{$^{2}$ Northern Robotics Laboratory, Universit\'e Laval, Canada; francois.pomerleau@ift.ulaval.ca}%
\thanks{$^{3}$ Irstea, Campus des Cezeaux, 63178 Aubi\`ere Cedex, France; christophe.debain@irstea.fr}%
}

\begin{document}

\maketitle
\thispagestyle{empty}
\pagestyle{empty}

\begin{abstract}
	In a context of autonomous robots, one of the most important tasks is to ensure the safety of the robot and its surrounding.
	The risk of navigation is usually said to be the probability of collision.
	This notion of risk is not well defined in the literature, especially when dealing with occupancy grids.
	The Bayesian occupancy grid is the most used method to deal with complex environments.
	However, this is not fitted to compute the risk along a path by its discrete nature.
	In this article, we present a new way to store the occupancy of the environment that allows the computation of risk along a given path.
	We then define the risk as the force of collision that would occur for a given obstacle.
	Using this framework, we are able to generate navigation paths ensuring the safety of the robot.
\end{abstract}

\section{INTRODUCTION} 
		Autonomous vehicles are nowadays more and more visible in our life.
		Most of them can be found in warehouses or on the road.
		As they evolve in a complex world, they need to assess the best choices to make regarding the possible obstacles.
		These choices are guided by the notion of risk: the robot must not harm others or itself.
		\citet{Fraichard2007} introduced this notion for known obstacles: the robot has not to collide with others to ensure its safety.

\begin{figure}[tbp]
				\centering
				\includegraphics[width=\linewidth]{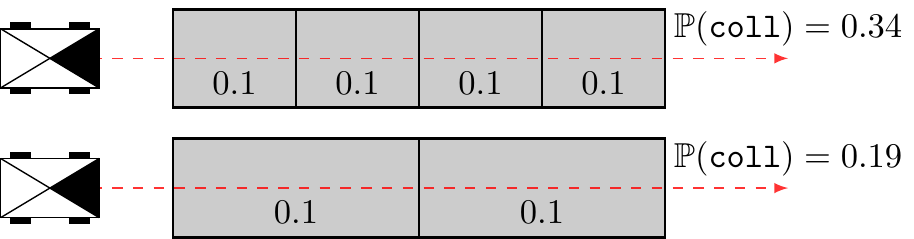}
				\caption{The robots (black boxes with their front represented as a filled triangle) want to cross an environment by following the dashed red line.
				The collision probability is uniform for the whole environment ($0.1$).
				The discretization size greatly influences the probability of collision, with the bottom scenario yielding to a safer path even though the underlying environment is the same.
				}
				\label{fig:intro}
				\vspace{-6mm}
		\end{figure}

		The most common way to store and deal with obstacles is the occupancy grids \cite{Elfes1989}.
		The map is discretized into a finite number of cells, where each cell stores the probability of occupancy.
		While navigating, the main concern is to ensure the safety of the robot and its surroundings.
		The most commonly used metric is the probability of collision, as the impact between the robot and another physical object is the main hazard \cite{Fraichard2007}. 
		In occupancy grid, we would be tempted to assess the probability of collision as the joint probability of colliding each cell.
		This simplicity hides a huge drawback that appears when computing this probability for two discretizations of the same map.
		\autoref{fig:intro} shows a robot wanting to cross an environment where the probability of occupancy is $0.1$.
		We discretized the environment with two different cell sizes.
		For the first one, we need to compute the probability that at least one cell is occupied, \ie $0.34$.
		For the second one, we only need to compute this probability over two cells, leading to $0.19$.
		We see that probabilities of collision for crossing the same part of the environment are completely different and dependent on the discretization size.
		Yet the grids store the same information using different discretizations, thus should give the same probability of collision.
		This problem can also be encountered while dealing with occupancy grids stored in quad-trees \cite{Kraetzschmar2004}.
		Indeed, the robot could decide to cross a large high-probability cell instead of ten small low-probability cells.

		We define the risk as a quantification of the danger encountered along a path.
		More precisely, this risk will be quantified as the force of collision the robot expects from a path.
		It is indeed more `risky' to hit a wall at high speed than at low-speed. 
		We propose in this article a novel method to compute the risk over a path.
		Our key contributions are
		\begin{itemize}
				\item A novel type of map, called Lambda-Field, specially conceived to allow path integrals over it;
				\item A mathematical formulation of the collision probability over a path; and
				\item A definition of the risk encountered over a path, specified as the expected force of collision along a path.
		\end{itemize}

		\autoref{sec:relatedWorks} presents a survey of the different methods to store the occupancy of the environment and the attempts to assess collision probabilities in these maps.
		\autoref{sec:theory} describes the theory of the Lambda-Field and shows the inherent application to risk assessment. 
		Finally, we give a few examples of Lambda-Field in \autoref{sec:experiment}.

\section{RELATED WORK} \label{sec:relatedWorks}
		In the context of path planning, the first step is to store the information of occupancy.
		Two methods have been proposed, which are the feature-based and metric-based maps.
		The first provides a list of all (possibly probabilist) obstacles in the environment.
		Although this is a very efficient way to store the occupancy, it is not easy to create a feature-based map from sensor data.
		Hence, feature-based maps are mainly created by humans and given as a prior to the robot \cite{Missiuro2006}.
		When the map is not available or the environment is too complex to be stored in a feature-based map (\eg a forest with unstructured obstacles), metric-based maps are used.
		In the simplest case, metric-based maps store the probability of occupancy for each position in the environment.
		\citet{Elfes1989} proposed the idea of tessellating the sensed environment and \citet{Coue2006} enhanced this idea, adding a bayesian layer. 
		Many variations of the Bayesian Occupancy Filter have been developed over the years, mainly adding dynamic obstacles in the grid.
		\citet{Saval-Calvo2017} wrote a review of the different Bayesian Occupancy Filter frameworks.
		Finally, \citet{OCallaghan2012} proposed a way to store the occupancy map without discretization, using gaussian process.
		This method allows to keep the dependence between cells in the grid which is not done in standard occupancy grid \cite{Elfes1989}.
	 	\citet{Ramos2016} developed an analog method using Hilbert Maps, overcoming the computational complexity of the gaussian process.

		Once a representation of the environment is available, the robot can start planning trajectories.
		Many of the popular methods use a binary representation of the environment, meaning that any point in the environment is either free or occupied.
		A review for such algorithms can be found in \cite{Tsardoulias2016}. 
		However, Bayesian occupancy grids are probabilist and do not give such binary information.
		The most common way to convert the bayesian grid into a binary grid is to apply a user-defined threshold \cite{Yang2013}:
		if the probability of occupancy is over this threshold, the cell is considered as occupied, otherwise it is free.
		Nevertheless, we loss a lot of information doing that, and some unstructured obstacles (\eg bushes) may become free after the conversion. 
		In the same fashion, \citet{Fulgenzi2007} chose to cluster the obstacles, leaving the unclustered space as free or occluded.
		The cost of the path is then simply the probability to collide with at least one cluster.
		It does, however, need to cluster the obstacles, which is not an easy task especially in unstructured environments.
		The above methods chose to consider a free space where no collision can happen: its computation can be difficult and noisy, leading to unpredicted collisions.
		We thus need a way to evaluate the cost of a path taking into account the probability of occupancy.
		Using Rapidly-exploring Random Tree (RRT), \citet{Fulgenzi2009} define the cost of a path as the joint probability of not having a collision in each node.
		It assumes that traveling between nodes is risk-free. 
		If we do not take this assumption, we fall back on the initial problem to compute a cost over a path.
		\citet{Gerkey2008} compute the cost of a path by summing the probability of occupancy of the cells the path crosses.
		This sum is then injected into a global cost function, taking into account other constraints like the speed or the distance to the objective, where each constraint has a user-defined coefficient.
		\citet{Francis2018} use the same idea for path planning in Hilbert Maps \cite{Ramos2016}.
		The drawback of this method is that the cost lacks physical meaning (as we sum probabilities):
		since this sum does not have any physical unit, its associated coefficient does not have one either, making its tuning non-intuitive for the user.
		Finally, \citet{Heiden2017} used the concept of product integral to compute the probability of collision over a path.
		It leads to a probability of collision, but this method has no physical meaning, which can lead to counter-intuitive probabilities.
		%
		Therefore, we propose a way to store the occupancy where the probability of collision over a path logically arises from the theory.

		However, the probability of collision may be not enough to quantify the safety of a path.
		The risk is often addressed in the context of known dynamics obstacles.
		In this configuration, the robot aims to avoid a configuration leading to a collision \cite{Fraichard2007}.
		This notion is extended by \citet{Althoff2010} with probabilistic obstacles.
		In the same fashion, the Time To Collision (TTC) \cite{Lee1976} is a very popular  metric of the risk.
		The TTC is useful in accident mitigation systems, but is not well fitted for long-term planning.
		\citet{Laugier2011} demonstrate its limitations, as, for instance, the TTC lacks context and sometimes leads to overestimate the risk.
		Regardless, these metrics only work for known obstacles and this information is not usually available in occupancy grids.
		\citet{Rummelhard2014} define the risk in a Bayesian occupancy grid as the probability to collide with a specific area, as well as the maximum probability of collision over the cells. 
		Nonetheless, these two risks have no physical meaning. 
		In our work, we thereupon define the risk as the expected force of collision on a given path.
\section{THEORETICAL FRAMEWORK} \label{sec:theory}
				The key concept of the Lambda-Field is its capability to assess the probability of collision inside a subset of the environment. 
				It relies on the mathematical theory of the Poisson Point Process. 
				This process counts the number of events which have happened given a certain period or area, depending on the mathematical space.
				In our case, we want to count the number of the event `collision' which could occur given a path (\ie a subset of $\mathbb{R}^2$). 

				For a positive scalar field $\lambda(x),x\in\mathbb{R}^2$, the probability to encounter at least one collision in a path $\mathcal{P}\subset \mathbb{R}^2$ is
				\begin{equation}
						\pr{ \mathtt{coll}|\mathcal{P} } = 1 - \expp{-\int_\mathcal{P}\lambda(x)\dif x}.
				\end{equation}
				Nonetheless, it is impossible to both compute and store the field $\lambda(x)$.
				Hence, we discretize our field into cells in a similar fashion to Bayesian occupancy grids.
				Under the assumption that the cells are small enough, the probability of collision can be approximated by
				\begin{equation}
					\pr{ \mathtt{coll}|\mathcal{P} } \approx 1 - \expp{-\Lambda(\mathcal{C})}, \hspace{3mm}\Lambda(\mathcal{C}) = \area\sum_{c_i\in\mathcal{C}}\lambda_i
						\label{eq:approxColl}
				\end{equation}
				for a path $\mathcal{P}$ crossing the cells $\mathcal{C}=\{c_0,\dots,c_N\}$, where each cell $c_i$ has an area of $\mathcal{A}$ and an associated lambda $\lambda_i$, which is the intensity of the cell. 
				The lambda can be seen as a measure of the density of the cell: the higher the lambda is, the most likely a collision will happen in this cell.

\begin{figure}[thb]
						\centering
						\includegraphics[width=\linewidth]{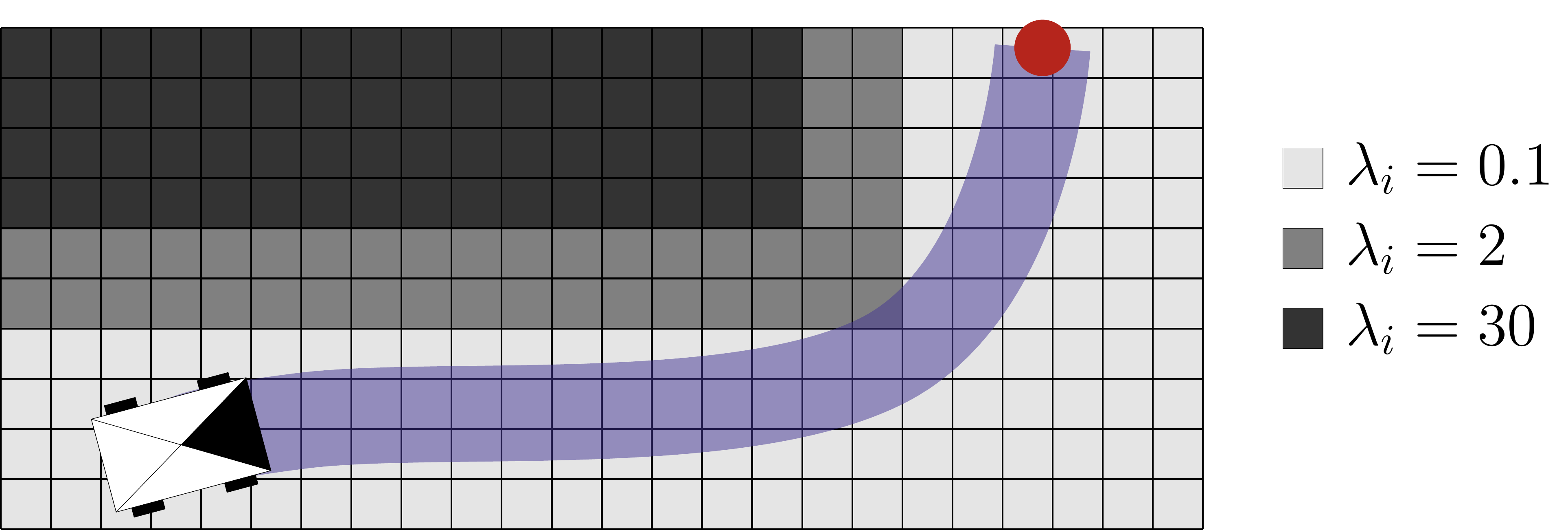}
						\caption{The robot (black boxes with its front represented as a filled triangle) wants to go to the position in red. In blue, the actual path the robot follows. Each cell has an area of $\SI{0.04}{\m\squared}$. Using \autoref{eq:approxColl}, the probability of collision in this path is $0.27$.}
						\label{fig:poissonpointprocess}
						\vspace{-2mm}
				\end{figure}

				Using this representation, we hereby see that the probability of collision is not dependent on the size of the cells.
				It is indeed the same to compute the probability of collision for crossing two cells of area $\area/2$ or one cell of area $\area$ for a constant $\lambda$.
				\autoref{fig:poissonpointprocess} gives an example of a path the robot might follow, as well as the underlying cells (of area $\SI{0.04}{\m\squared}$) it crosses. 
				The robots crosses $58$ cells with $\lambda_i=0.1$ and one cell with $\lambda_i=2$.
				Using \autoref{eq:approxColl}, the probability of collision is evaluated at $0.27$. 

		\vspace{-4mm}	
		\subsection{Computation of the field}
\begin{figure}[thb]
						\centering
						\includegraphics[width=\linewidth]{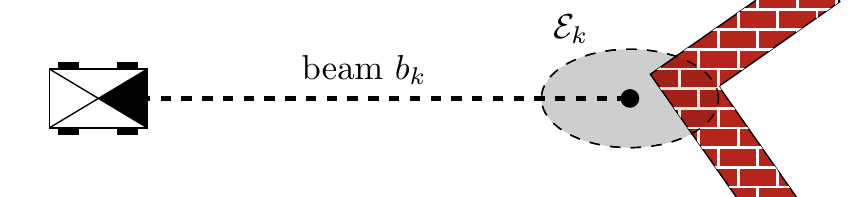}
						\caption{The robot (black boxes with its front represented as a filled triangle) measures an obstacle. The obstacle is in the area $\mathcal{E}_k$ centered on the measurement.}
						\label{fig:errorRegion}
						\vspace{-2mm}
				\end{figure}

			As we established a new approach to represent the occupancy of an environment, we need to develop a way to compute dynamically the lambdas. 
			We assume that the robot is equipped with a lidar sensor, which gives us a list of cells crossed by beams without collision, and another list of cells where the beams collided.
			Also, we represent the uncertainty of the sensor in a fashion that differs from the standard forward models \cite{Elfes1989}, given three variables: 
			\begin{itemize}
				\item $\mathcal{E}_k$, the region of error of the lidar for the beam $b_k$ with its associated area $e_k$. 
					It represents the accuracy of the sensor, and can be of any shape: it means that the true position of the obstacle is within the region $\mathcal{E}_k$, centered on the obstacle measurement from the beam $b_k$.
					We set the shape and size of $\mathcal{E}_k$ constant for every measurements. We have thus $e_k=e$ for all beams.
					\autoref{fig:errorRegion} gives an example of such a region.
				\item $p_m$, the probability of rightfully read `miss' for a cell (\ie the cell is not in the region of error $\mathcal{E}_k$). The quantity $1-p_m$ gives the probability to read `miss' for a cell that should be in the region of error $\mathcal{E}_k$.
				\item $p_h$, the probability of rightfully read `hit' for a cell (\ie the cell is in the region of error $\mathcal{E}_k$). The quantity $1-p_h$ gives the probability to read `hit' for a cell that is empty. The probability $p_h$ is for example much lower when the sensor is in the rain, as many readings comes from raindrops.
			\end{itemize}
			

			Using this sensor model, we construct the Lambda-Field in the following manner.
			We want to find the combination of $\lambda=\{\lambda_i\}$ that maximizes the expectation of the $K$ beams the lidar has shot since the beginning.
			For each lidar beam $b_k$, the beam crossed without collision the cells $c_m\in \mathcal{M}_k$ and hit an obstacle contained in the cells $c_h\in\mathcal{E}_k$.
			The log-likelihood of the beam $b_k$ is
			\begin{multline}
			\mathcal{L}(b_k|\lambda) = \ln\left[ \expp{-\Lambda(\mathcal{M}_k)}\left( 1-\expp{-\Lambda(\mathcal{E}_k)}\right) \right].
			\end{multline} 

			The log-likelihood of $K$ lidar beams is then 
				\begin{equation}
						\begin{aligned}
							&\mathcal{L}(\{b_k\}_{0:K-1}|\lambda) = \sum_{k=0}^{K-1} \mathcal{L}(b_k|\lambda) \\
													&= \sum_{k=0}^{K-1} \left[ -\Lambda(\mathcal{M}_k) + \ln\left(1- \expp{-\Lambda(\mathcal{E}_k)}\right)\right].
						\end{aligned}
				\end{equation}
				
				We want to maximize this quantity, hence nullify its derivative since the function is concave.

				In order to find a closed-form, we approximate the derivative with the assumption that the variation of lambda inside the region of error of the lidar is small enough to be negligible. Thus, for each $\lambda_i\in\mathcal{E}_k$ we have
				\begin{equation}
					\area\sum_{c_h\in\mathcal{E}_k}\lambda_h \approx e\lambda_i.
				\end{equation}
				Using this approximation, the derivative is 
				\begin{equation}
						\frac{\partial \mathcal{L}(\{b_k\}_{0:K-1}|\lambda)}{\partial \lambda_i} \approx -m_i\cdot\mathcal{A} + h_i\frac{\mathcal{A}}{\exp(e\lambda_i)-1},
				\end{equation}
				where $h_i$ is the number of times the cell $c_i$ has been counted as `hit' (\ie was in the region of error of the sensor) and $m_i$ is the number of times the cell $c_i$ has been counted as `free'.
				
				We finally find the zero of the derivative, leading to
				\begin{equation}
						\Leftrightarrow \lambda_i = \frac{1}{e}\ln\left(1+\frac{h_i}{m_i}\right).
						\label{eq:lambda}
				\end{equation}

				This closed-form allows a very fast computation of the lambda field.
				We also see that the formula is independent on the size of the cells.
				
				\subsection{Confidence intervals}
				In the same way as \citet{Agha-mohammadi2016}, we define the notion of confidence over the values in the Lambda-Field.
				Indeed, the robot should not be as confident over a certain path if the cells have been read one time or one hundred times.
				For each cell $c_i$, we seek the bounds $\lambda_L$ and $\lambda_U$ such that
				\begin{equation}
					\begin{aligned}
						&\mathbb{P}(\lambda_L \le \lambda_i \le \lambda_U) \ge 95\%\\
						\Leftrightarrow &\mathbb{P}(\lambda_L \le \frac{1}{e}\ln\left(1+\frac{h_i}{m_i}\right) \le \lambda_U) \ge 95\%.
					\end{aligned}
				\end{equation}
				Using the relation $h_i = M - m_i$ where $M$ is the number of times the cell has been measured, we can rewrite the above equation as
				\vspace{-2mm}
				\begin{equation}
					\mathbb{P}(K_L \le h_i \le K_U) \ge 95\%,
				\end{equation}
				such that
				\begin{equation}
					\begin{aligned}
						\lambda_L &= \frac{1}{e}\ln\left(\frac{K_L}{M-K_L}+1\right), \\
						\lambda_U &= \frac{1}{e}\ln\left(\frac{K_U}{M-K_U}+1\right).
					\end{aligned}
					\label{eq:KLKU}
				\end{equation}
%
				The quantity $h_i$ can be seen as a sum of two binomial variables:
				\begin{equation}
					\begin{aligned}
						h_i &= \bar{h} + (m_i-\bar{m}) &\bar{h}\sim \mathbb{B}(p_h,h_i) \\
							&= \bar{h} + \bar{w}  &\bar{w} \sim \mathbb{B}(1-p_m,m_i),
					\end{aligned}
				\end{equation}
				where $\bar{h}$ (resp. $\bar{m}$) is the number of times the sensor rightfully read a `hit' out of the $h_i$ trials (resp. read a `miss' out of the $m_i$ trials). 
				The quantity $(m_i-\bar{m})$ is hence the number of times the sensor wrongfully read `hit' instead of `miss'.

				The distribution of $h_i$ is not binomial but a Poisson binomial distribution with poor behaviors in terms of computation. 
				Since the Poisson binomial distribution satisfies the Lyapunov central limit theorem, we can approximate its distribution with a Gaussian distribution of same mean and variance:
				\vspace{-4mm}
				\begin{equation}
					\begin{aligned}
						\mu &= h_ip_h+m_i(1-p_m), \\
						\sigma^2 &= h_i(1-p_h)p_h + m_i(1-p_m)p_m.
					\end{aligned}
				\end{equation}
				We can then have the bounds at 95\%, with
				\begin{equation}
						\begin{aligned}
							K_L &= \max(\mu-1.96\sigma, 0) ,\\
							K_U &= \min(\mu+1.96\sigma, M). 
						\end{aligned}
				\vspace{0mm}
				\end{equation}
				The bounds $\lambda_L$ and $\lambda_U$ are then retrieved from $K_L$ and $K_U$ using \autoref{eq:KLKU}.

\begin{figure}[thp]
					\vspace{-3mm}
					\centering
					\includegraphics[width=\linewidth]{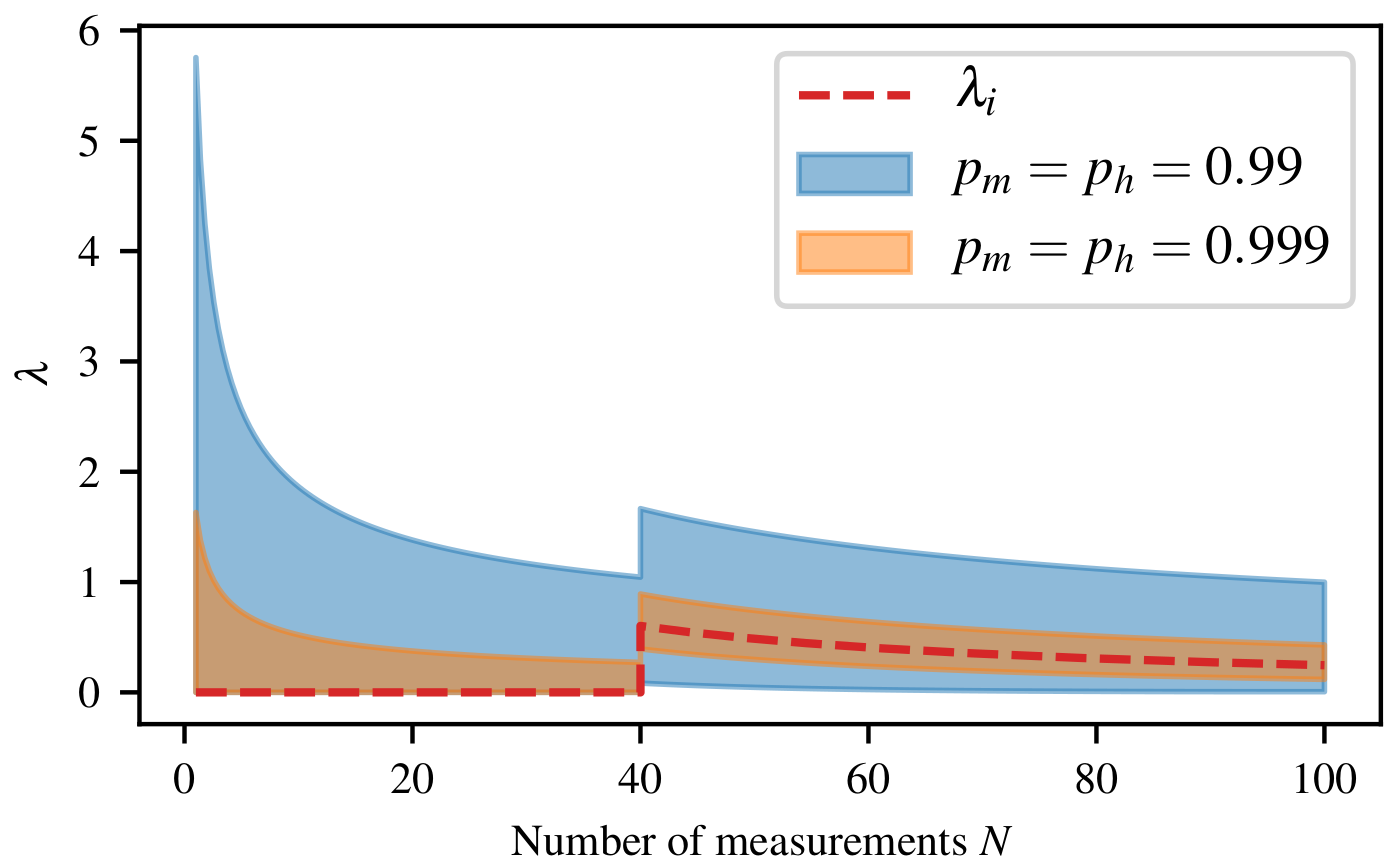} 
					\caption{Convergence of the confidence intervals for a free cell $c_i$. At the fortieth measurement, the sensor misreads the cell and returns a `hit'. The confidence interval grows around the expected $\lambda_i$ before re-converging.}
					\label{fig:intervals}
					\vspace{-5mm}
				\end{figure}

				\autoref{fig:intervals} shows an example of behavior of the confidence interval for different confidences. 
				The lidar measures an empty cell $c_i$. The confidence interval quickly decreases as the number of readings `miss' increases.
				At the fortieth measurement, the lidar misreads and returns a `hit' for the cell.
				The confidence interval grows around the expected lambda computed with \autoref{eq:lambda} before re-converging.

				\vspace{0mm}
			\subsection{Risk assessment}
\begin{figure}[thp]
				\centering
				\includegraphics[width=\linewidth]{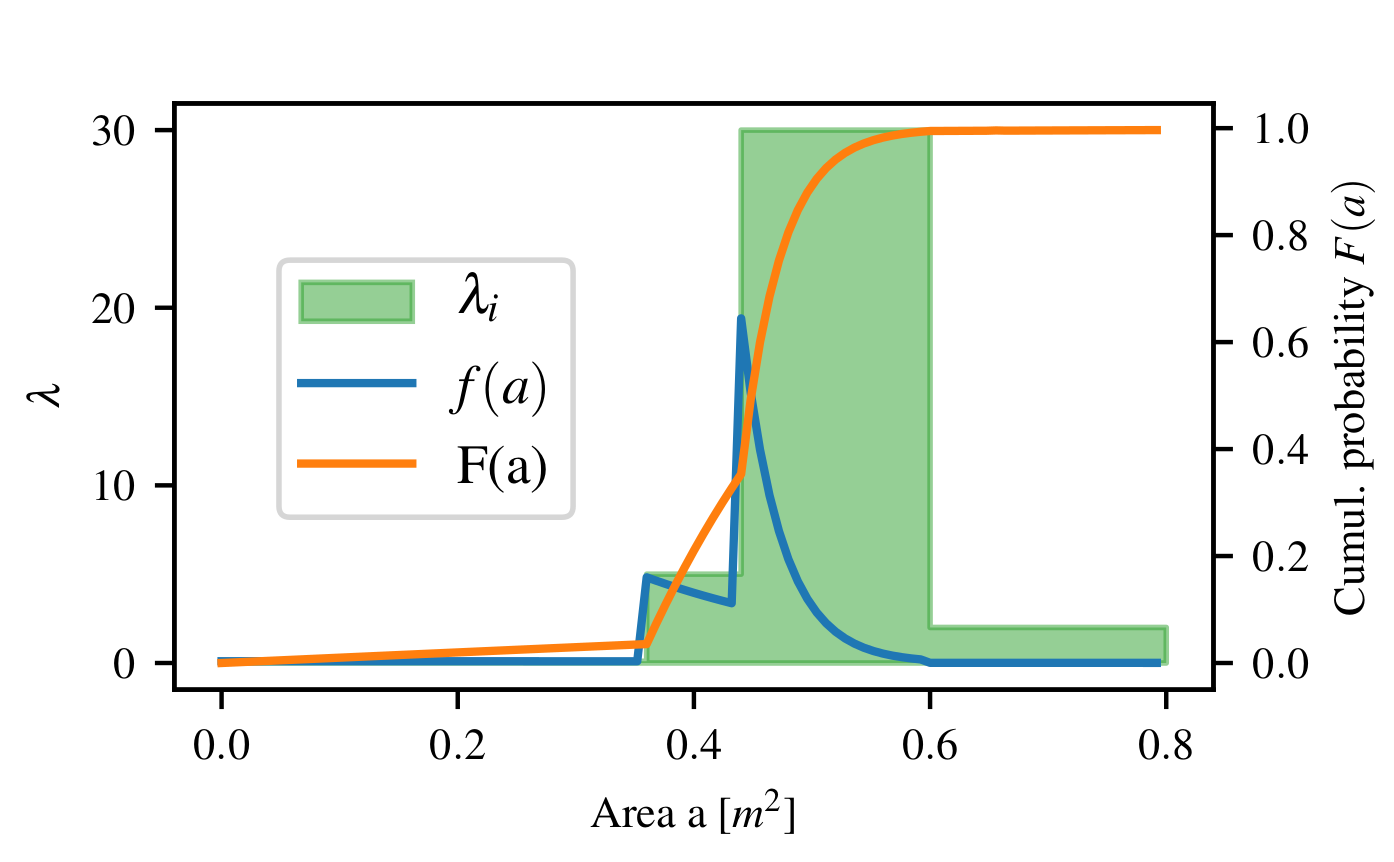}
				\caption{Example of lambda field the robot crosses (in green), with the associated probability distribution $f(a)$ (in blue) and cumulative probability distribution $F(a)$ (in orange).} 
				\label{fig:probaFunction}
		\end{figure}

				As said before, the motivation of the Lambda-Fields is its ability to compute path integral, hence a risk along a path.
				For a path $\mathcal{P}$ crossing the cells $\mathcal{C}=\{c_i\}_{0:N}$ in order, the probability distribution function over the Lambda-Field is
		\begin{equation}
				f(a) = \expp{n\mathcal{A}\lambda_n-\mathcal{A}\sum_{i=0}^{n-1}\lambda_i} \cdot \lambda_n\cdot\expp{-a\lambda_n},
		\end{equation}
		where $n=\floor{a/\mathcal{A}}$ and $\floor{\cdot}$ is the floor function. The variable $a$ denotes the area the robot has already crossed. 
		\autoref{fig:probaFunction} shows an example of the probability density for a given path on a Lambda-Field: when the robot goes through high-lambda cells, the cumulative distribution probability quickly increases to one.
		This can be easily proved as integrating $f(a)$ over a certain path $\mathcal{P}$ crossing the cells $\mathcal{C}$ gives the probability of encountering at least one collision:
		\begin{equation}
				\pr{\mathtt{coll}|\mathcal{P}} = \int_\mathcal{P} f(a)\dif a = 1 - \expp{-\Lambda(\mathcal{C})}.
		\end{equation}
		We can then define the expectation of a risk function $r(\cdot)$ over the path:
		\vspace{-2mm}
		\begin{equation}
				\mathbb{E}[r(X)] = \int_\mathcal{P} f(a) r(a) \dif a
		\end{equation}
		The random variable $X$ denotes the position (\ie area) at which the first event `collision' occurs.
		Most of the time, the cells are small enough to assume that the function $r(\cdot)$ is constant inside each cell.
		Using this assumption, we simplify the above equation to
		\begin{equation}
			\mathbb{E}[r(X)] = \sum_{i=0}^{N} r(\area i)\expp{-\area \sum_{j=0}^{i-1}\lambda_j}\left(1-\expp{-\area \lambda_i}\right),
			\label{eq:risk}
		\end{equation}
		for a path $\mathcal{P}$ going through the cells $\{c_i\}_{0:N}$.

	\begin{figure*}[tbp] 
		\begin{subfigure}[t]{0.3\textwidth}
			\centering
			\includegraphics[width=\linewidth]{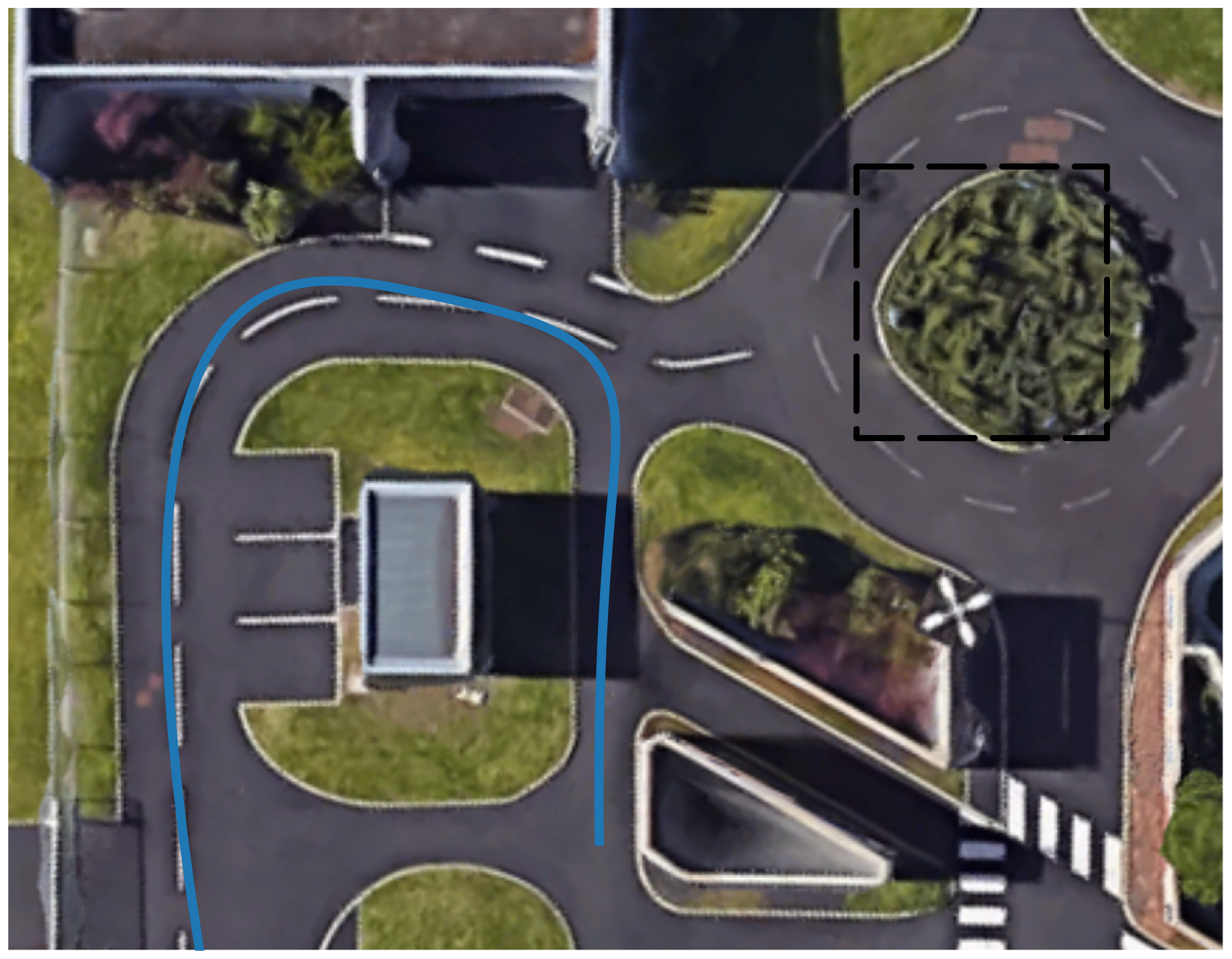}
			\caption{}
		\end{subfigure}%
		~
		\begin{subfigure}[t]{0.7\textwidth}
			\centering
			\includegraphics[width=\linewidth]{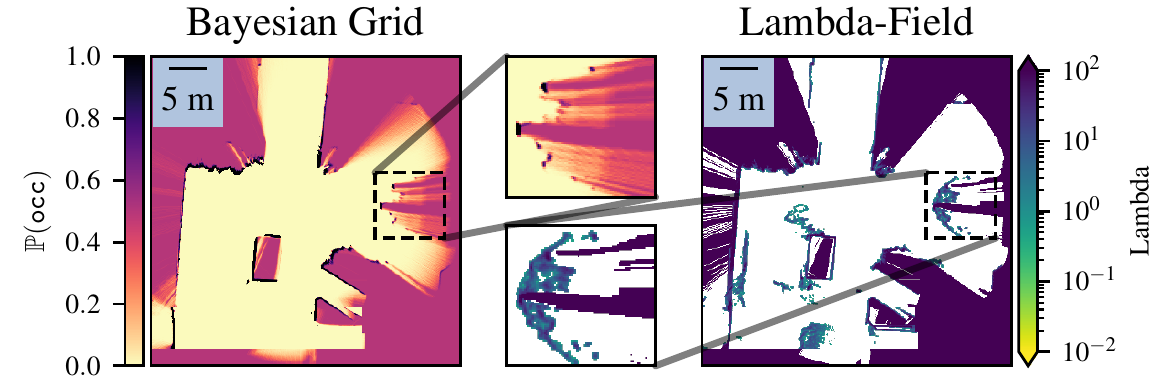}
			\caption{}
		\end{subfigure}
		\caption{(a) Aerial View of the mapped environment, with the robot path in blue and the roundabout in dashed black (b) \textit{Left:} Bayesian occupancy grid \textit{Right:} Lambda-Field. The Lambda-Field is better suited to store the occupancy of unstructured obstacles where the Bayesian Occupancy Filter may over-converges, especially for the roundabout (dashed black).}
		\label{fig:compareMaps}
		\vspace{-1em}
	\end{figure*}

	\begin{figure}[tbp]
			\centering
			\includegraphics[width=\linewidth]{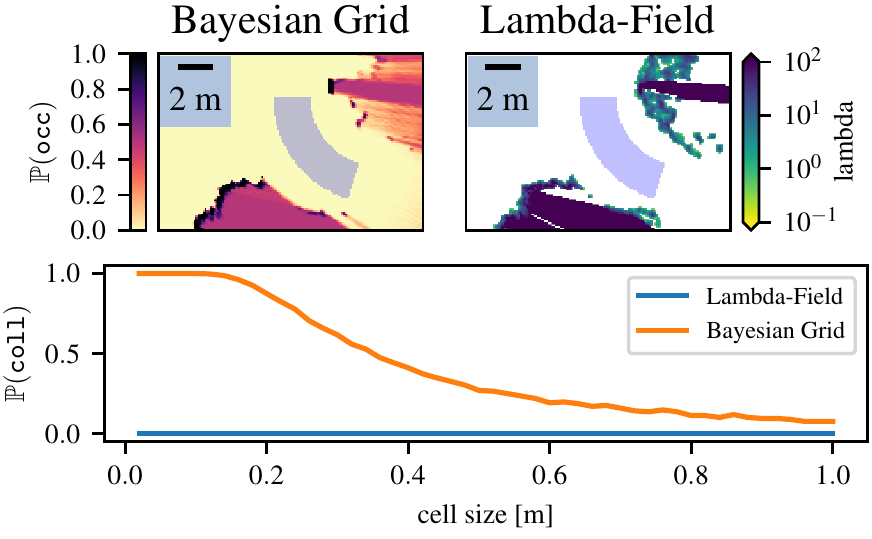}
			\caption{Comparison of the probability of collision for a free-obstacle path in Bayesian occupancy grid and Lambda-Field. The natural way to compute the probability of collision of the path (in light blue) in Bayesian occupancy grid is dependent of the tessellation size, as opposed to the Lambda-Field which is not.}
			\label{fig:compareRisk}
	\end{figure}
		In our case, we are interested in the risk of a certain path.
		In other words, the function $r(\cdot)$ will be the risk encountered at each instant or position.

		We chose to model the risk as the force of collision (\ie loss of momentum) if the collision occurred at the area $a$, which is
		\begin{equation}
			r(a) = m_R\cdot v(a),
			\label{eq:qttmvt}
		\end{equation}
		where $m_R$ is the mass of the robot, and $v(a)$ is its velocity at the area $a$. 
		One can note that it is quite easy to convert $a$ into the curvilinear abscissa, which is far more convenient to link to the speed. 
		For a robot of width $w$ which has crossed an area $a$, its curvilinear abscissa $s$ equals to
		\begin{equation}
			s=\frac{a}{w}.
		\end{equation}

		This metric, fast to process, assumes that every obstacle the robot might encounter has an infinite mass. 
		It means that if the robot collides with an obstacle, the resulting collision would lead the robot to stop (\ie losing a momentum of $m_R\cdot v$). 

		One can note that setting $r(a)=1$ leads to the probability of collision, given by \autoref{eq:approxColl}.
		More complicated metrics can of course be developed.
		For instance, the angle of collision can be taken into account to better quantify the loss of momentum, if this information is available.


\section{EXPERIMENTATIONS} \label{sec:experiment}
	
%

	In order to demonstrate the effectiveness of our framework, we implemented our method onto a mobile robot.
	The robot navigates through a real-world environment with structured and unstructured obstacles.
	\autoref{fig:compareMaps} (a) shows an aerial view of the terrain.
	The robot is equipped with a lidar Sick LMS-151 which gives range measurements.
	While navigating, a Lambda-Field as well as a Bayesian occupancy grid is created.
	\autoref{fig:compareMaps} (b) gives an example of maps.
	We discretized the field into cells of size $10\times\SI{10}{\cm}$.
	The sensor has the probability of $p_h=0.99$ and $p_m=0.9999$ to read the right information.
	The value of $p_m$ is intentionally really large since it is almost impossible for a lidar beam to entirely miss an obstacle.
	However, it is far more possible that the lidar returns a hit for a free region (when the beam hits a raindrop, for example).
	We also choose to model the region of error of the lidar as a disk of area $\SI{0.04}{\m\squared}$.
	Among the differences between the two maps, one can see that the roundabout on the right side of the map is not very well represented.
	The roundabout is indeed made of small bushes where the lidar beams can go through quite easily.
	The Bayesian occupancy grid discarded most of the roundabout.
	Indeed, the Bayesian Occupancy Filter has to converge either to `occupied' or `free' and some of the cells misconverge due to of the sparsity of the obstacle.
	On the other hand, the Lambda-Field keeps more information as it stores a measure of density which does not need to converge to an extremum.

	In the meantime, the robot had to follow a pedestrian while ensuring its safety.
	We implemented the path-planning method presented by \cite{Gerkey2008}.
	Every second, the robot samples trajectories, parametrized as a velocity $v$ and a rotational velocity $\omega$ applied for one second.
	Then, it chooses the best trajectory, which is the one that stays the closest to the path (from a global path planning algorithm between the robot and the pedestrian).
	For each trajectory, we first process its associated upper limit risk using \autoref{eq:risk} and the upper bound of the lambdas. 
	All the trajectories that present an upper limit risk above the maximum risk allowed are discarded.
	If none of the trajectories is acceptable, the robot chooses not to move as it is its only admissible decision. 
	In our case, we chose that the maximum risk is $\SI{1}{\kg\m\per\s}$, meaning that we are sure at $95$ percents that the robot will not encounter collision with a force above this maximum.

	\autoref{fig:compareRisk} presents a very simple comparison of the probability of collision inferred from a Bayesian occupancy grid and a Lambda-Field. 
	The robot assesses the probability of collision for a path going around the roundabout (in light blue).
	The most intuitive way to assess the probability of collision in Bayesian occupancy grids is to compute the probability of not colliding any cell in the path.
	It leads to a probability of collision that highly depends on the cell size, since a smaller cell size means more cells not to collide to.
	In the other hand, the expected probability of collision in the Lambda-Field does not depend on the cell size. 
	Indeed, the overall integration of the lambdas does not depend on the tessellation size.
	\citet{Heiden2017} also defined a collision probability that does not depend on the tessellation size.
	Their formulation lacks physical meaning, leading a probability that can only be used to compare with other probabilities and not prove that a path is truly safe.

\section{CONCLUSION}
	In this article, we present a novel representation of occupancy of the environment, called Lambda-Field.
	We first derived a way to fill the map, as well as confidence intervals over these values.
	This representation specifically allows the computation of path integrals, giving a natural way to assess the probability of collision.
	Using the Lambda-Field, we are able to compute the risk of collision over a path, defined as the force of collision.
	We finally tested our framework in a real environment, showing the usefulness of our method to assess the probability of collision as well as the risk along a path.

	Future work will estimate the mass and the angle of collision to better estimate the risk.
	We will also test our framework in more challenging environments like snowy forests.
	Finally, dynamic obstacles will be addressed as most urban scenarios are not static.


\section*{ACKNOWLEDGMENT}
This work was funded by grants from the French program `investissement d'avenir' managed by the National Research Agency (ANR), the European Commission (Auvergne FEDER funds) and the `R\'egion Auvergne' in the framework of the LabEx IMobS 3  (ANR-10-LABX-16-01).


\renewcommand*{\bibfont}{\small}
\printbibliography

\end{document}